%% file: neurips_2025.tex
\documentclass{article}

\usepackage[final]{neurips_2025}

\usepackage[utf8]{inputenc} 
\usepackage[T1]{fontenc}    
\usepackage{hyperref}       
\usepackage{url}            
\usepackage{booktabs}       
\usepackage{amsfonts}       
\usepackage{nicefrac}       
\usepackage{microtype}      
\usepackage{xcolor}         

\usepackage{graphicx} 
\usepackage{amsmath}

\usepackage{fvextra}
\DefineVerbatimEnvironment{verbatim}{Verbatim}{breaklines,breakanywhere=true}

\usepackage{pifont}    

\definecolor{ForestGreen}{RGB}{34,139,34} 
\newcommand{\diffdown}[1]{\textcolor{red}{$\downarrow$}\,#1}
\newcommand{\diffup}[1]{\textcolor{ForestGreen}{$\uparrow$}\,#1}
\usepackage{comment}
\usepackage{tikz}            
\usetikzlibrary{calc}

\title{Too Late to Recall: Explaining the Two-Hop Problem in Multimodal Knowledge Retrieval}

\usepackage{hyperref}       
\usepackage{fontawesome5}   

\title{Too Late to Recall: Explaining the Two-Hop Problem in Multimodal Knowledge Retrieval}

\author{%
\begin{minipage}{\linewidth}
Constantin Venhoff\textsuperscript{1,4}\quad
Ashkan Khakzar\textsuperscript{1,4}\quad
Sonia Joseph\textsuperscript{2,3}\quad
Philip Torr\textsuperscript{1}\quad
Neel Nanda
\end{minipage}
\vspace{0.3cm}
\and
\vspace{0.5cm}
\textsuperscript{1}University of Oxford \quad
\textsuperscript{2}McGill University \quad
\textsuperscript{3}Meta \quad
\textsuperscript{4}MATS\\[4pt]
\faGithub\;\href{https://github.com/cvenhoff/vlm-two-hop}{cvenhoff/vlm-two-hop}
}

\begin{document}

\maketitle

\input{sec/0_abstract}    
\input{sec/1_intro}
\input{sec/2_method}
\input{sec/related_work}

\input{sec/3_conclusion}

{\small
\bibliographystyle{unsrtnat}
\bibliography{main}
}








\appendix

\input{sec/X_suppl}
\clearpage

\end{document}

%% file: sec/0_abstract.tex
\begin{abstract}
Training vision language models (VLMs) aims to align visual representations from a vision encoder with the textual representations of a pretrained large language model (LLM). However, many VLMs exhibit reduced factual recall performance compared to their LLM backbones, raising the question of how effective multimodal fine-tuning is at extending existing mechanisms within the LLM to visual inputs.
We argue that factual recall based on visual inputs requires VLMs to solve a two-hop problem: (1) forming entity representations from visual inputs, and (2) recalling associated factual knowledge based on these entity representations. By benchmarking 14 VLMs with various architectures (LLaVA, Native, Cross-Attention), sizes (7B-124B parameters), and training setups on factual recall tasks against their original LLM backbone models, we find that 11 of 14 models exhibit factual recall degradation. We select three models with high and two models with low performance degradation, and use attribution patching, activation patching, and probing to show that degraded VLMs struggle to use the existing factual recall circuit of their LLM backbone, because they resolve the first hop too late in the computation. In contrast, high-performing VLMs resolve entity representations early enough to reuse the existing factual recall mechanism.
Finally, we demonstrate two methods to recover performance: patching entity representations from the LLM backbone into the VLM, and prompting with chain-of-thought reasoning. Our results highlight that the speed of early entity resolution critically determines how effective VLMs are in using preexisting LLM mechanisms. More broadly, our work illustrates how mechanistic analysis can explain and unveil systematic failures in multimodal alignment.
\end{abstract}

%% file: sec/1_intro.tex
\section{Introduction}
\begin{figure}
\centering
\includegraphics[width=\linewidth]{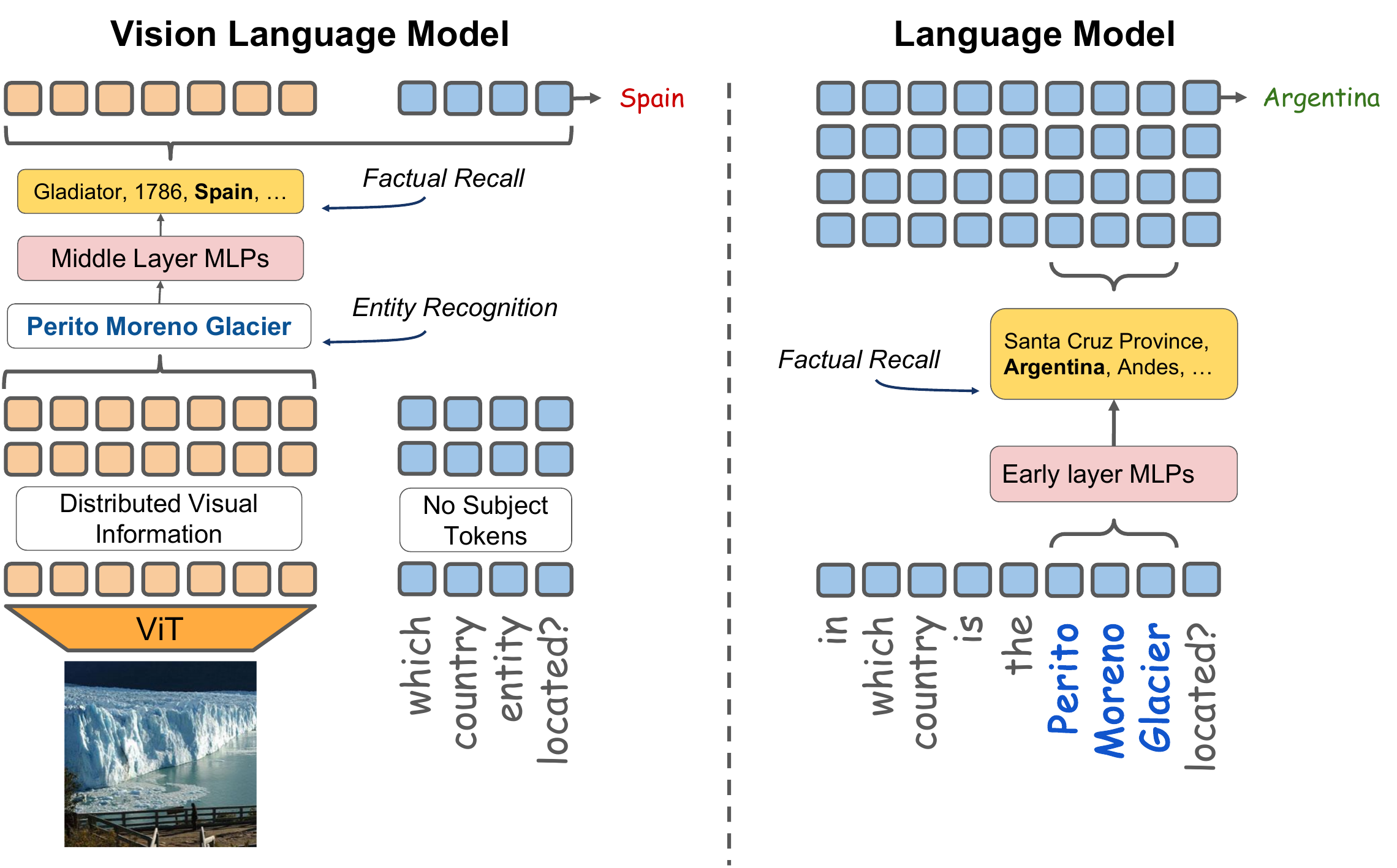}
\caption{\textbf{Factual Recall in VLMs vs. LLMs – Illustration of the Two-Hop Problem.}
This figure compares how a Vision Language Model (VLM, left) and a text-only Language Model (LLM, right) perform a factual recall task. The VLM receives an image of the Perito Moreno Glacier and a question: “In which country is the entity located?” The image is processed by a Vision Transformer (ViT), producing distributed visual embeddings that do not align with the LLM’s pretrained token space. The entity (“Perito Moreno Glacier”) is only recognized in the middle layers, bypassing the early-layer MLPs responsible for factual recall and resulting in an incorrect answer.
In contrast, the LLM is given the full question, with the subject tokens “Perito Moreno Glacier” available. This enables early-layer MLPs to access factual knowledge (“Argentina”) and produce the correct answer.
The comparison highlights the core issue: VLMs must first infer the subject before retrieving facts, but because recognition occurs late, it cannot engage early factual recall mechanisms. This “two-hop” problem leads to degraded factual accuracy in VLMs, even when visual recognition succeeds.
}
\label{fig:exp1}
\end{figure}

Vision–Language Models (VLMs) achieve strong multimodal task performance by integrating vision transformers (ViTs) with large language models (LLMs) via adapter mechanisms \cite{liu2023visual,lin2024vila,steiner2024paligemma2familyversatile,liu2024improved,cocchi2025llava,touvron2023llama, dosovitskiy2021imageworth16x16words}. These adapters project visual representations from the ViT into the representation space of the backbone LLM, enabling text-based reasoning over visual inputs. However, how exactly preexisting mechanisms in the backbone LLM adapt to visual inputs, and the resulting implications or failure modes, remain severely understudied.

For example, previous studies have found that LLaVA-1.5-7B~\citep{liu2024improved} performs worse on factual recall tasks than its LLM backbone \cite{cohen2024performance}. Explaining this requires understanding how visual information flows through VLMs and how factual recall mechanisms function in their LLM backbones. Prior work on factual recall mechanisms identified early-layer MLPs as a crucial component that reads in subject token representations and then produces subject-related factual representations \cite{meng2022locating,chughtai2024summing}. Separate work on VLM information flow, however, shows that visual projections are not aligned with the backbone LLM's token space and instead only gradually align with textual representations deeper in the LLM \cite{venhoff2025how,wu2025the,masry2025alignvlm}.

Combining these observations, we hypothesize that VLMs fail to adapt to the preexisting factual recall circuit of their LLM backbone because visual representations in early layers are misaligned with the backbone LLM's token space and therefore "skip over" early-layer MLPs. Factual recall in VLMs can therefore be viewed as a two-hop problem: the VLM must first construct robust entity representations from visual inputs before it can access the LLM's preexisting factual knowledge.

To investigate this, we benchmark 14 VLMs across various architectures (LLaVA, Native, Cross-Attention), model sizes (7B–124B parameters), and training setups on factual recall tasks, comparing their performance to their original LLM backbone models. We find that 11 out of 14 models exhibit factual recall degradation. We then conduct a comparative analysis of well- and poorly-performing VLMs using attribution patching, activation patching, and linear probing. We find that VLMs with degraded performance rely on different sublayers for factual recall than their LLM backbones, while high-performing VLMs reuse the same components. In degraded VLMs, visual entity representations emerge too late in the backbone LLM's forward pass, bypassing early-layer MLPs of the factual recall mechanism, whereas well-aligned models exhibit early resolution of such representations. By patching early-layer MLP outputs from the backbone LLMs into degraded VLMs, we can recover factual recall performance, providing strong causal evidence for our hypothesis. Finally, we explore chain-of-thought prompting as a mitigation strategy, finding varying but promising improvements, particularly for larger VLMs.

Our findings highlight that successful multimodal alignment requires more than representational compatibility: it depends on integrating visual information into the functional circuits of the LLM backbone.

%% file: sec/2_method.tex
\section{Benchmarking Factual Recall in Vision Language Models}
\label{sec:benchmark}

\begin{table*}[t]
  \caption{Factual-recall accuracy of VLMs vs.\ their text-only LLM backbones on 1000 questions.}
  \label{tab:factual-recall-final}
  \centering
  \begin{tabular}{l l l ccc}
    \toprule
    \textbf{VLM Model} & \textbf{LLM Model} & \textbf{Architecture} &
    \textbf{LLM (\%)} & \textbf{VLM (\%)} & \textbf{$\Delta$ (\%)}\\
    \midrule
    LLaVA-MORE-8B        & Llama-3.1-8B-it    & Adapter       & 41 & 23 & \diffdown{43.9}\\
    LLaVA-NEXT-8B        & Llama-3-8B-it      & Adapter       & 41 & 24 & \diffdown{41.5}\\
    LLaVA-1.5-7B         & Llama-2-7B-chat    & Adapter       & 30 & 19 & \diffdown{36.7}\\
    LLaVA-1.5-13B        & Llama-2-13B-chat   & Adapter       & 44 & 28 & \diffdown{36.4}\\
    Pixtral-Large-124B   & Mistral-Large-2    & Adapter       & 68 & 56 & \diffdown{17.6}\\
    Pixtral-12B          & Mistral-NeMo       & Adapter       & 41 & 36 & \diffdown{12.2}\\
    Qwen2.5-VL-7B-it    & Qwen2.5-7B-it     & Adapter       & 30 & 28 & \diffdown{6.7}\\
    Qwen2.5-VL-72B-it   & Qwen2.5-72B-it    & Adapter       & 49 & 53 & \diffup{-8.2}\\
    \midrule
    Llama-4-Maverick     & —                  & Native        & 75 & 71 & \diffdown{5.3}\\
    Gemini-2.0-Flash     & —                  & Native        & 66 & 63 & \diffdown{4.5}\\
    Gemma-3-27B-it       & —                  & Native        & 46 & 46 & 0.0\\
    Gemma-3-12B-it       & —                  & Native        & 40 & 40 & 0.0\\
    GPT-4o               & —                  & Native  & 77 & 73 & \diffdown{5.2}\\
    \midrule
    Llama-3.2-Vision-11B & Llama-3.1-8B-it    & Cross-attn    & 36 & 31 & \diffdown{13.9}\\
    \bottomrule
  \end{tabular}
\end{table*}

To systematically evaluate factual‑recall degradation in vision–language models (VLMs), we introduce a benchmark that directly compares each VLM with its original LLM backbone model. By feeding equivalent information to both systems, the benchmark isolates factual‑recall ability from other confounding factors.

\subsection{Benchmark Design}
Our benchmark comprises 15000 multimodal factual‑recall questions.  We sample images from the Wikipedia‑based Image–Text (WIT) dataset \citep{wit} and use GPT‑4.1 to generate entity‑specific factual prompts (e.g.\ “Who invented the entity shown in the image?”).  GPT‑4.1 was chosen for its strong factual accuracy on the WildHallucinations benchmark \citep{zhao2024wildhallucinations}.  Appendix~\ref{sec:dataset-pipeline} details the data‑generation pipeline, including entity verification and question construction.

During evaluation, a VLM is first asked to identify the main entity in each image. If the entity is misidentified, the sample is discarded to avoid conflating recognition errors with factual retrieval. We conduct a brief case study into rejected examples in Appendix~\ref{app:exclusion}, to ensure there is no structural bias in the images discarded. For correctly identified entities, we compare the VLM’s answer with that of its language‑only backbone, ensuring a controlled, like‑for‑like comparison (for prompts see Appendix~\ref{app:prompts}).

We test a broad range of architectures, including \emph{Adapter‑based} VLMs, where a pretrained LLM is connected to a vision transformer (ViT) through lightweight adapters, \emph{Native} VLMs, which are trained end‑to‑end with multimodal data from the beginning, and \emph{Cross-Attention} VLMs where the LLM backbone uses cross-attention blocks to attend to representations of the ViT.  Each VLM/backbone pair answers factual recall questions until 1000 valid samples are answered from the 15000‑question pool.

\subsection{Benchmark Results}
Table~\ref{tab:factual-recall-final} summarizes the results: 11 out of 14 VLMs exhibit lower factual‑recall accuracy than their language‑only backbones. Adapter‑based- and Cross-Attention-models suffer the steepest drop. Degradation persists even for very large models such as Pixtral‑Large‑124B.  In general, Adapter‑based models degrade more than Native VLMs, with the notable exception of Qwen2.5‑VL.  We hypothesize that Qwen2.5‑VL’s extensive multimodal fine-tuning on over 4 trillion tokens mitigates factual recall degradation, which may also explain why the Qwen2.5‑VL‑72B‑it variant slightly \emph{outperforms} its backbone.

In summary, almost all tested VLMs show pronounced factual‑recall degradation, especially Adapter-based VLMs, while Native models such as Llama‑4‑Maverick and GPT‑4o degrade less. Qwen2.5‑VL’s success suggests that massive multimodal fine-tuning can enable the VLM to utilize the factual knowledge of its backbone LLM.

\subsection{Selecting Models for Comparative Analysis}
In the following sections we describe several experiments to understand the drivers of factual recall degradation in VLMs. We select three Adapter-based models with very high degradation, LLaVA-MORE-8B \citep{cocchi2025llava}, LLaVA-1.5-7B \citep{liu2023visual}, and LLaVA-1.5-13B \citep{liu2024improved}. Additionally, we select two models with very low degradation, Gemma-3-12B-it \citep{gemmateam2025gemma3technicalreport} to study Native models, and Qwen2.5-VL-7B-it \citep{bai2025qwen25vltechnicalreport} to understand the impact of massive multimodal fine-tuning on Adapter-based models.

\section{Comparing Factual Recall Mechanisms}
\label{sec:attribution}
\begin{figure*}
\centering
\includegraphics[width=\linewidth]{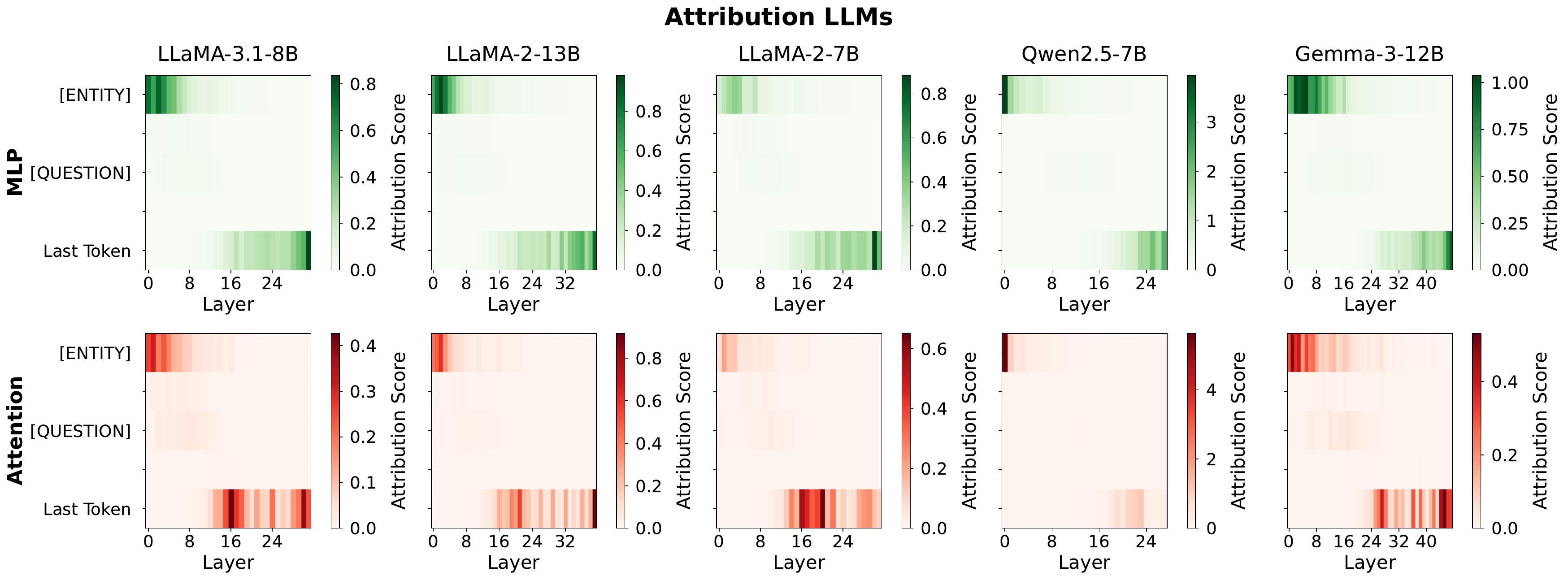}
\caption{Attribution scores of each MLP and Attention sublayer for the original LLM backbone models. Higher values indicate higher causal relevance for factual recall.}
\label{fig:exp3_llm}
\end{figure*}
\begin{figure*}
\centering
\includegraphics[width=\linewidth]{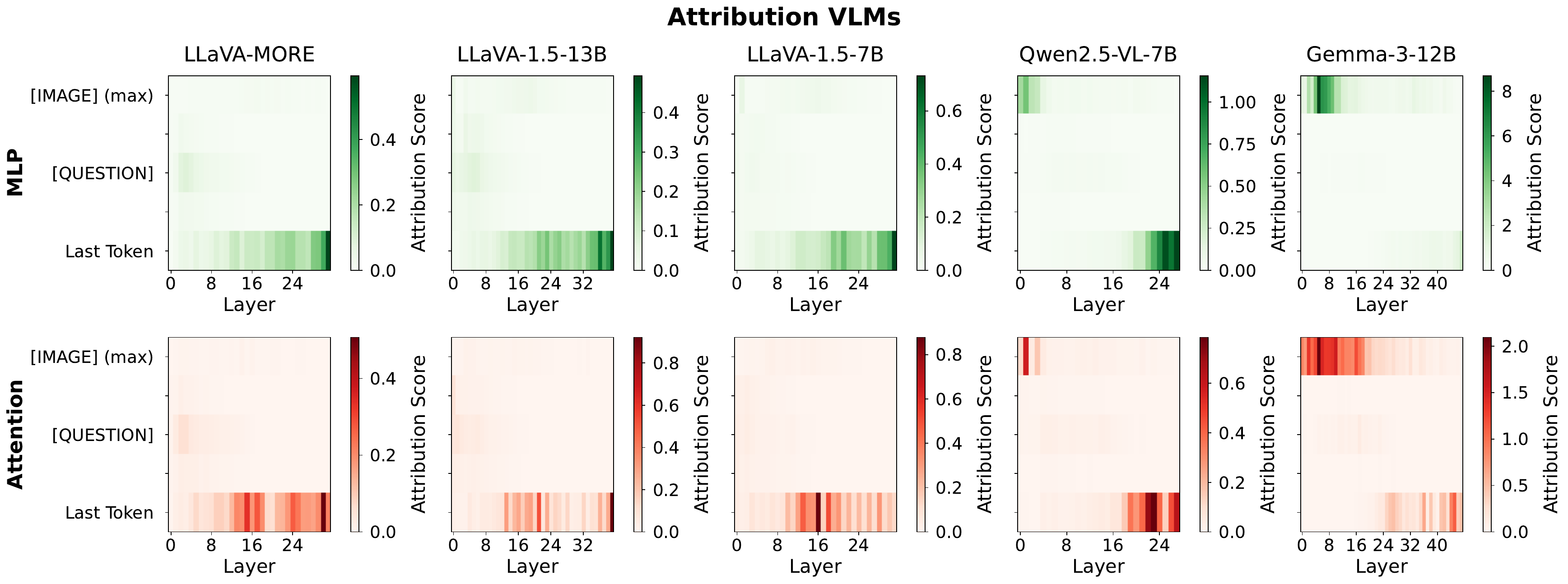}
\caption{Attribution scores of each MLP and Attention sublayer for the VLM models. Higher values indicate higher causal relevance for factual recall.}
\label{fig:exp3_vlm}
\end{figure*}

In this section we aim to find the sublayers the original LLM backbone model uses to perform factual recall, and compare them against the sublayers that the respective VLM model uses. 

\subsection{Attribution Patching}
To determine which MLP- and Attention-sublayers in the LLM and VLM contribute most to factual recall, we employ an attribution patching methodology inspired by \citeauthor{nanda2023attribution} and \citeauthor{meng2022locating}. First, we sample 100 correctly answered examples from the benchmark dataset for each VLM and their original LLM backbone model. Then we use the following steps to compute the attribution scores:
\begin{enumerate}
  \item \textbf{Targeted corruption}:
  Let $\mathcal{S}$ be the token span necessary to identify the entity (LLM: entity text tokens; VLM: image tokens).
  Compute $\sigma_{\text{embed}}$ as the standard deviation of input embeddings over $\mathcal{S}$ across the examples of the attribution dataset.
  Perform a \textit{corrupted run}, by injecting Gaussian noise in the embeddings across $\mathcal{S}$:
  \[
  H^{(0)}_{\mathcal{S}} \leftarrow H^{(0)}_{\mathcal{S}} + \varepsilon,\quad
  \varepsilon \sim \mathcal{N}\!\big(0,\;\alpha\,\sigma_{\text{embed}}\big),
  \]
  where $\alpha$ is a noise multiplier. We ablate $\alpha$ for each model over the values $\{1,2,\dots,9,10\}$ and measure the corruption effect via the KL divergence between the clean predicted token distribution, and the corrupted predicted token distribution. More details on how we choose $\alpha$ can be found in Appendix~\ref{app:noise}.

  \item \textbf{Clean/corrupted passes and KL objective}:
  Run the model without the corruption (clean) and with the corruption (corrupted) to obtain next-token prediction logits $(z_{\text{clean}}, z_{\text{corr}})$ and define
  \[
  L \;=\; D_{\mathrm{KL}}\!\big(\mathrm{softmax}(z_{\text{clean}})\,\Vert\,\mathrm{softmax}(z_{\text{corr}})\big).
  \]

  \item \textbf{Per-layer branch activations and gradients}:
  For each layer $\ell$ and branch $b\in\{\mathrm{mlp},\mathrm{attn}\}$, cache clean and corrupted outputs
  $H^{\text{clean}}_{\ell,b},\,H^{\text{corr}}_{\ell,b}$ and backpropagate to obtain
  \[
  G_{\ell,b} \;=\; \frac{\partial L}{\partial H^{\text{corr}}_{\ell,b}}.
  \]

  \item \textbf{Grad×Delta attribution and aggregation}:
  For token $t$ and hidden dimension $d$,
  \[
  a_{\ell,b}(t)
  \;=\;
  \Big|\sum_{d} G_{\ell,b}(t,d)\,\big(H^{\text{clean}}_{\ell,b}(t,d)-H^{\text{corr}}_{\ell,b}(t,d)\big)\Big|.
  \]
\end{enumerate}
Figure~\ref{fig:exp3_llm} shows the attribution scores for the LLMs and Figure~\ref{fig:exp3_vlm} shows the attribution scores for the VLMs averaged over the 100 correct factual recall examples from the benchmark dataset, aggregated across different token positions (entity/image, question, and last token positions).

\subsection{Attribution Patching Results}
We find the same two sites across all LLMs tested: the early-layer MLP- and Attention-sublayers across the entity tokens, and the middle-to-late layer MLP- and Attention-sublayers across the last token positions. This is consistent with findings from \cite{meng2022locating}. For all LLaVA-style VLMs we find that there is only the mid-to-late layer site across the last token position active, indicating that the model is not able to use the same pathway as the LLM backbone. For Gemma-3-12B and Qwen2.5-VL-7B however, we find the same two sites, showing that the model uses a similar pathway, whether the entity is given in token space, or as an image.

\section{Causal Analysis via Activation Patching}
\label{sec:patching}
\begin{figure}
\centering
\includegraphics[width=\textwidth]{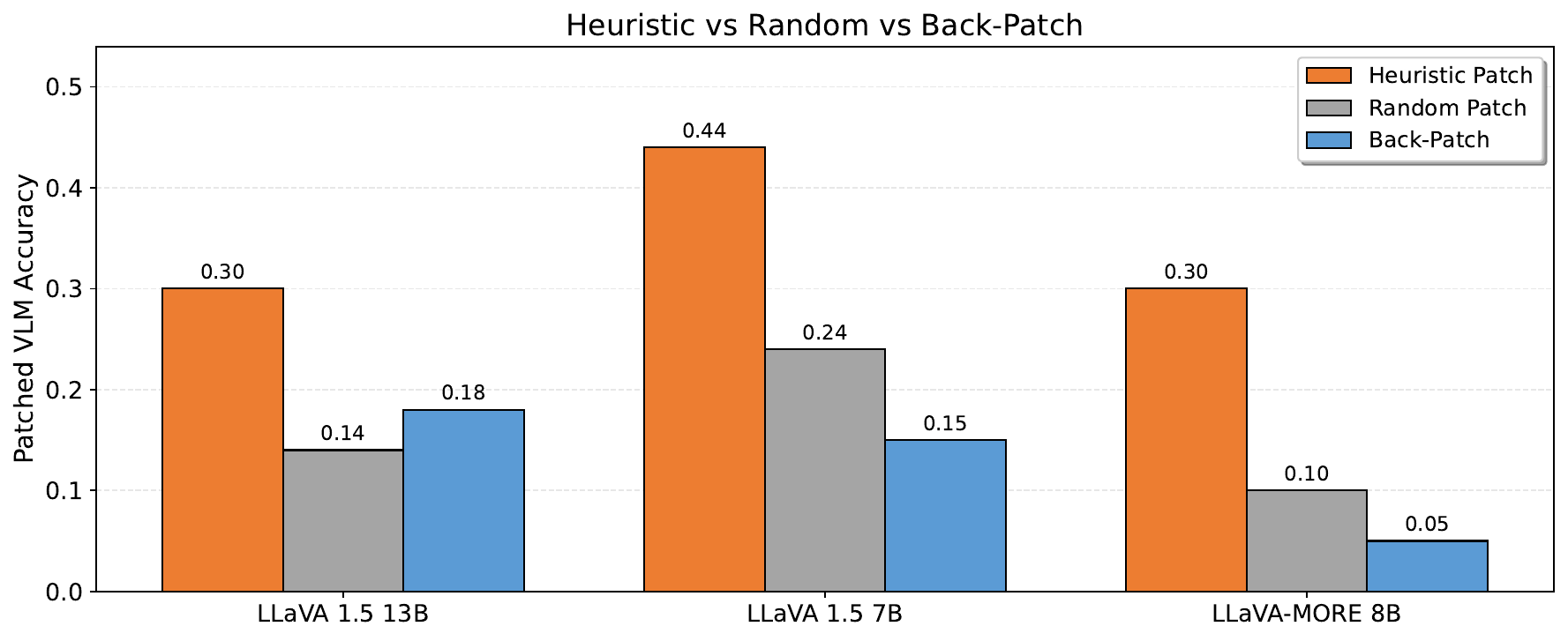}
\caption{Factual recall performance recovery for LLaVA-1.5-7B, LLaVA-1.5-13B, and LLaVA-MORE models when MLP outputs across entity tokens from corresponding LLM backbones are patched into early VLM layers. We test only on examples where the VLM was originally wrong, so the y-axis directly shows the recovered performance gap. We compare our heuristic patching approach against a random baseline (randomly selecting patching positions), and back-patching.}
\label{fig:exp3-2}
\end{figure}

In Section~\ref{sec:attribution}, we found that for the three LLaVA models, the LLM backbone uses early-layer MLPs at the entity token positions to perform factual recall, while the LLaVA models that are trained on top of these LLMs are not able to do that. We hypothesize therefore, that these early-layer MLPs are a key breaking point in the factual recall mechanism of the LLaVA models. To test this, we patch MLP outputs from the LLM backbones of the three LLaVA models into their forward pass to see whether this yields relevant performance recovery, which would support that the LLaVA models are not able to utilize the factual recall circuit of their LLM backbone properly. 

\subsection{Heuristic Patching}
We propose \textit{heuristic patching} to determine which token position in the VLM we patch the backbone LLM's MLP outputs (as the VLM doesn't have corresponding entity token positions). Heuristic patching uses 4 steps:

\begin{enumerate}
  \item \textbf{Caching MLP outputs}:
  We define a set of source layers $\mathcal{L}$. For each layer $\ell \in \mathcal{L}$, cache the LLM's MLP outputs over the entity tokens.

  \item \textbf{Define patching objective}:
  We define an objective, that quantifies whether patching the cached LLM MLP outputs into a given VLM token span, from a given start position $p$, increases the likelihood of the VLM to change its generated answer to the LLM answer. To do this, we obtain clean next-token predictions from both models. If the next-tokens predictions of the LLM and VLM differ, we use a logit-difference objective that increases the LLM-chosen token and decreases the VLM-chosen token. If they match, we use a KL objective over the LLM's top 10 tokens, renormalized. We treat both as a \emph{signed} objective so that higher is better; we do not take absolute values.

  \item \textbf{Position-wise attribution}:
  For every candidate token start position $p$ in the VLM, compute an attribution score by backpropagating the chosen objective to the VLM MLP outputs and taking a Grad$\times\Delta$, where $\Delta$ is the activation difference between LLM MLP activations and the current MLP activations across the VLM slice with start positions $p$. Average the scores across the slice token positions, keep the sign, and sum the scores across all layers in $\mathcal{L}$ for each token start position $p$ to obtain single scores $s(p)$.

  \item \textbf{Selection and patching}:
  Choose the top start position with the highest $s(p)$. We patch \emph{all} layers in $\mathcal{L}$ from the selected start token position during VLM generation and record whether the VLM now predicts the correct answer.
\end{enumerate}

\noindent \textbf{Baselines.} The results of heuristic patching are hard to interpret in isolation, as 1) LLaVA models are fine-tuned end-to-end, hence the LLM backbone in the LLaVA model is not the same as the original LLM backbone model, and 2) there is no designated token position to patch into. Therefore it is unclear what accuracy to expect or consider \textit{good}. Thus, we compare against (i) \emph{Random patching}, which uses the same patching rule as heuristic patching, but chooses the start token positions $p$ uniformly at random, and (ii) \emph{Back-patching}, a two-pass procedure both used in prior work to address two-hop problems in LLMs \citep{biran2024hopping} and VLMs \citep{nikankin2025taskdifferentcircuitsdisentangling}, that copies entire layer outputs across the image-tokens from a set of source layers $S$ to a set of (earlier) destination layers $D$.
Concretely, we use:
\begin{itemize}
  \item LLaVA-MORE (LLaMA-3.1-8B backbone): $S{=}[6..16]$, $D{=}[1..11]$.
  \item LLaVA-1.5-7B (LLaMA-2-7B backbone): $S{=}[6..16]$, $D{=}[1..11]$.
  \item LLaVA-1.5-13B (LLaMA-2-13B backbone): $S{=}[27..37]$, $D{=}[13..23]$.
\end{itemize}
We select for each VLM/LLM pair a set of 100 factual recall questions from our dataset, which were incorrectly answered by the VLM and correctly answered by the corresponding LLM. Then we run heuristic patching with different layer choices (see Appendix~\ref{app:patching} for details), and compare against the two baseline methods.

\subsection{Patching Results}
Results are shown in Figure~\ref{fig:exp3-2}. We find that on average heuristic patching recovers 35\% of the factual recall accuracy difference between the VLM and its language-only counterpart, while back-patching only recovers 13\%, and random patching recovers 16\%. We therefore show that the early-layer MLP outputs used in the LLM's factual recall circuit are indeed a core missing piece in the degraded factual recall circuit used by the LLaVA-style VLMs.

\section{Probing for the Emergence of Visual Entity Representations}
\label{sec:probing}
\begin{figure}
\centering
\includegraphics[width=\linewidth]{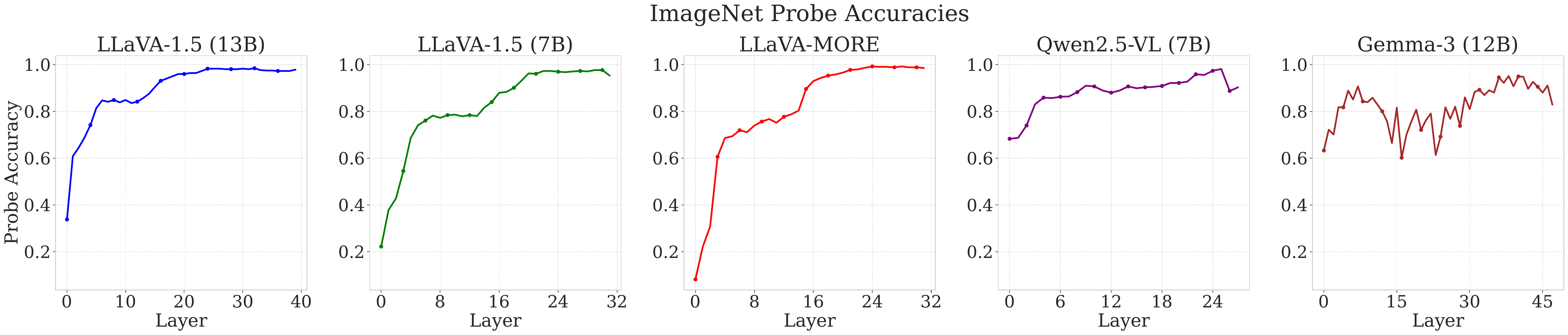}
\caption{Accuracy of linear probes trained on residual-stream representations at each transformer layer of LLaVA-1.5-7B, LLaVA-1.5-13B, LLaVA-MORE, Gemma-3-12B, and Qwen2.5-VL-7B measured on ImageNet-100 entity prediction. The three LLaVA-style models exhibit a consistent pattern: probe accuracy remains poor in early layers and rises sharply between middle-to-late layers. Gemma-3-12B and Qwen2.5-VL-7B on the other hand show consistently high probe accuracies.}
\label{fig:exp4}
\end{figure}

Our previous experiments demonstrate that LLaVA-style VLMs show degraded factual recall performance, because they are not able to utilize the factual recall circuit of their LLM backbone correctly. 
The Native model Gemma-3-12B and the Adapter-based model Qwen2.5-VL-7B, trained on billions of image-text pair tokens, on the other hand, are seemingly able to use the same circuit, whether they are prompted with an image of an entity or the entity in text tokens.

We know already from prior work that the visual representations passed to the LLM backbone in LLaVA-style models are not aligned with the token space (reproduced in Appendix~\ref{app:token_space}). Therefore, in order to use the same circuit as the LLM backbone, VLMs need to produce robust entity representations in early layers of the LLM backbone.
We hypothesize that the reason for the degraded performance of the LLaVA-style models is that entity representations emerge too late in the forward pass, compared to Gemma-3-12B and Qwen2.5-VL-7B which might be able to produce such representations much earlier due to native pretraining and massive multimodal fine-tuning respectively.

\subsection{Linear Probing for Visual Entity Representations}
\label{subsec:linear_probing}
To systematically assess where visual entity representations emerge within the LLM backbone of VLMs, we employ a probing methodology. Specifically, we train layer-wise linear classifiers (probes) on the residual stream outputs of each transformer layer to predict visual entities depicted in input images.

\paragraph{Experimental Setup:}  
We use a subset of ImageNet-100 \citep{imagenet}, restricted to 50 of the 100 classes, as our evaluation dataset, due to its controlled set of 50 classes and multiple images per entity, ensuring reliable estimation of representational capacity. We randomly select 2500 images and generate for each image a factual recall question, using a similar methodology as in Section~\ref{sec:benchmark}. We provide GPT-4.1 with the entity class, and prompt for a concise factual recall question. We keep track of the last few generated questions and prompts to produce a different type of question, to ensure sufficient question diversity. At each transformer layer, we train an independent linear probe on the extracted and averaged residual-stream representations over the question token positions, to predict the correct entity label. We use a 20\%/80\% train-test split to evaluate the probes. 

\paragraph{Results and Analysis:}  
Figure~\ref{fig:exp4} depicts the accuracy of the linear probes across layers for LLaVA-1.5-7B, LLaVA-1.5-13B, LLaVA-MORE, Gemma-3-12B, and Qwen2.5-VL-7B. The results reveal a clear trend for the LLaVA-style models: linear representations capable of reliably encoding visual entity information do not emerge until the middle-to-late layers. Prior to these middle-to-late layers, probe accuracy remains poor, indicating that early layers do not encode robust entity representations. Gemma-3-12B and Qwen2.5-VL-7B on the other hand show consistently high probe accuracies across all layers (with small variance), indicating that these models are capable of producing robust entity representations from visual inputs in the LLM backbone immediately.

These findings reveal the core challenge of the \textbf{two-hop problem} in multimodal knowledge retrieval: the model must first form a robust entity representation (first hop) before being able to access the latent knowledge of the LLM backbone encoded in early-layer MLPs (second hop). However, since entity representations only emerge in deeper layers in LLaVA-style models, they bypass the early-layer MLPs, causing factual recall degradation.

\section{Reasoning as a Potential Mitigation}
\label{sec:reasoning}
\begin{table*}[t]
  \caption{Factual-recall accuracy using \textit{Chain-of-Thought (CoT) prompting}. 
  Green arrows denote improvement compared to the base (non-CoT) runs.}
  \label{tab:factual-recall-cotonly}
  \centering
  \begin{tabular}{l l l cc}
    \toprule
    \textbf{VLM Model} & \textbf{LLM Model} & \textbf{Architecture} &
    \textbf{LLM (CoT, \%)} & \textbf{VLM (CoT, \%)}\\
    \midrule
    LLaVA-MORE-8B        & Llama-3.1-8B-it    & Adapter       & 46\,\diffup{+5.0}  & 28\,\diffup{+5.0} \\
    LLaVA-NEXT-8B        & Llama-3-8B-it      & Adapter       & 45\,\diffup{+4.0}  & 30\,\diffup{+6.0} \\
    LLaVA-1.5-7B         & Llama-2-7B-chat    & Adapter       & 35\,\diffup{+5.0}  & 18\,\diffdown{–1.0} \\
    LLaVA-1.5-13B        & Llama-2-13B-chat   & Adapter       & 45\,\diffup{+1.0}  & 37\,\diffup{+9.0} \\
    Pixtral-Large-124B   & Mistral-Large-2    & Adapter       & 72\,\diffup{+4.0}  & 71\,\diffup{+15.0} \\
    Pixtral-12B          & Mistral-NeMo       & Adapter       & 45\,\diffup{+4.0}  & 48\,\diffup{+12.0} \\
    Qwen2.5-VL-7B-it    & Qwen2.5-7B-it     & Adapter       & 35\,\diffup{+5.0}  & 43\,\diffup{+15.0} \\
    Qwen2.5-VL-72B-it   & Qwen2.5-72B-it    & Adapter       & 52\,\diffup{+3.0}  & 56\,\diffup{+3.0} \\
    \midrule
    Llama-4-Maverick     & —                  & Native        & 79\,\diffup{+4.0}  & 77\,\diffup{+6.0} \\
    Gemini-2.0-Flash     & —                  & Native        & 70\,\diffup{+4.0}  & 70\,\diffup{+7.0} \\
    Gemma-3-27B-it       & —                  & Native        & 53\,\diffup{+7.0}  & 57\,\diffup{+11.0} \\
    Gemma-3-12B-it       & —                  & Native        & 41\,\diffup{+1.0}  & 46\,\diffup{+6.0} \\
    GPT-4o               & —                  & Native        & 79\,\diffup{+2.0}  & 79\,\diffup{+6.0} \\
    \midrule
    Llama-3.2-Vision-11B & Llama-3.1-8B-it    & Cross-attn    & 42\,\diffup{+6.0}  & 40\,\diffup{+9.0} \\
    \bottomrule
  \end{tabular}
\end{table*}

Previous experiments have shown that the LLaVA-style VLMs are fundamentally constrained in accessing the factual knowledge of their LLM backbone. The only mitigation seems to be either native pretraining, or massive multimodal fine-tuning, which both require orders of magnitude more compute and data than LLaVA-style models. Therefore, we experimented with another, simpler, mitigation strategy. The VLMs could use inference-time compute, to describe the visual entities relevant for factual recall in text space, and then answer the question. In this setup, the VLM might be able to access the factual recall knowledge of its LLM backbone again, as it produces token representations related to visual entities and therefore doesn't rely on pure visual representations anymore.

\subsection{VLM Chain of Thought Prompting}
\label{subsec:cot}

\paragraph{Experimental Setup:}  
To test this, we alter the prompting template in the benchmark and include a chain of thought prompt (see Appendix~\ref{app:prompts} for prompt). Then we re-run the benchmark on the same data for each model and record the increase in factual recall accuracy.

\paragraph{Results and Analysis:}  
Table~\ref{tab:factual-recall-cotonly} shows the chain of thought benchmark accuracies for each model, alongside the increase in accuracy compared to the original accuracies without chain of thought. We find that for most models the chain of thought leads to a larger increase for the VLM compared to the LLM backbones, supporting our hypotheses. For Pixtral-12B and Pixtral-Large-124B we see that the performance gap is fully closed, while for the LLaVA-style models we see an increasing gain with the model size, with LLaVA-1.5-13B closing over half of its performance gap. Albeit having a smaller performance gap to begin with, we see that Gemini-2.0-Flash, GPT-4o, and LLama-4-Maverick close their respective performance gaps when using chain of thought prompting.

In summary, the effect of chain of thought prompting seems to be very model dependent (e.g. LLaVA-1.5-7B even \textit{loses} a percentage of accuracy, as chain of thought prompting often leads to less structured responses); however, models with substantial reasoning capabilities are able to close a significant fraction or even the full performance gap. We therefore think that reasoning is a promising approach for future research to improve factual recall performance of VLMs for which only limited data and compute resources are available.

%% file: sec/related_work.tex
\section{Related Work}

\label{sec:intro}
We review prior research on Vision Language Models (VLMs) and their handling of visual and textual representations. We first cover methods for aligning visual features with language embeddings and their effect on multimodal reasoning. We then summarize findings on factual recall in unimodal and multimodal models, highlighting evidence of recall degradation in VLMs.

\subsection{Vision Language Models}
VLMs integrate visual and textual information by mapping both modalities into a shared embedding space processed by a pretrained LLM \cite{liu2023visual}. A key challenge is ensuring that visual features retain critical information (e.g., entities) when projected into the token space \cite{masry2025alignvlm}.
The standard approach uses an image encoder and a projector (typically an MLP) to map visual data into the LLM’s embedding space \cite{liu2023visual, liu2024llavanext, chen2024internvl}. However, this mapping often fails to achieve meaningful alignment, limiting VLM performance on multimodal tasks \cite{masry2025alignvlm, lin2024vila}. To improve this, several works propose alternatives that refine the mapping process or introduce cross-modal layers deeper in the network \cite{alayrac2022flamingo, yan2024tg, li2024visual, upadhyay2023probvlm}. While these methods enhance interaction between modalities, they add significant computational cost.

\subsection{Multimodal Mechanistic Interpretability}
Recent interpretability research has explored vision language models through various approaches. \cite{chen2023interpreting}, \cite{gandelsman2023interpreting}, \cite{joseph2025steeringclipsvisiontransformer}, and  \cite{joseph2025prismaopensourcetoolkit} analyzed the vision encoder CLIP, identifying visual features. \cite{schwettmann2023multimodal} examined multimodal neurons in transformer models that were solely trained on text data. \cite{jiang2024interpreting} investigated VLM responses to hallucinated versus real objects, and showed how to edit their internal representations to mitigate hallucinations. Work by \cite{neo2024towards} investigated information flow within VLMs by projecting visual representations from intermediate layers onto language vocabulary, identifying gradual alignment with text vocabulary. \cite{naghashyar2025towards} use sparse autoencoders to show how visual features emerge during multimodal fine-tuning in LLaVA. 

\subsection{Factual Recall in LLMs and VLMs}
LLMs implement important mechanisms of factual recall in early-layer MLPs \cite{chughtai2024summing, meng2022locating, geva2023dissecting}. This layer localization has been linked to their struggles with multi-hop reasoning, which requires resolving intermediate entities before answering the full query \cite{biran2024hopping, sakarvadia2023memory, yang2024large}. For example, answering "Who is the mother of the inventor of transistors?" requires first identifying the inventor. \cite{biran2024hopping} suggests early layers resolve these "bridge entities," leaving later layers with insufficient capacity for the follow-up reasoning \cite{chughtai2024summing}. \cite{cohen2024performance} reports that LLaVA-1.5-7B struggles with factual recall when processing images. Their patching experiments reveal that visual tokens stop influencing model behavior beyond middle layers, implying that visual processing finishes before later layers handle factual reasoning. They hypothesize that factual recall failures occur due to insufficient remaining layers after visual processing.

%% file: sec/3_conclusion.tex
\section{Discussion and Conclusion}
\label{sec:discussion_conclusion}
Despite the growing capabilities of Vision Language Models (VLMs), their ability to reliably ground outputs in factual knowledge remains brittle. This is especially concerning in high-stakes applications, where factual hallucination carries real-world risks. In this work, we identify a structural failure mode in LLaVA-style VLMs: a systematic degradation of factual recall caused by insufficient early layer alignment of textual- and visual representations in the LLM backbone of the VLM.

Our findings show that such misalignment causes VLMs to use different sublayer-components than the original LLM backbone model to perform factual recall. This restricts them from accessing the full factual knowledge of their backbone LLM. We confirm this hypothesis using attribution patching, activation patching and probing experiments. We show that Native VLMs and adapter-based VLMs with massive multimodal fine-tuning do not exhibit the same degradation, as they are able to form consistent entity representations in early layers of the LLM backbone, regardless of the input modality.

These findings expose fundamental limitations in current LLaVA-style VLMs that are trained with limited data and compute resources. We conduct promising early experiments with chain of thought prompting to utilize inference-time compute to mitigate the problem, suggesting an avenue for future research. Additionally, future work should look into better, more data efficient, adapter mechanisms and multimodal alignment techniques to integrate visual representation earlier into the LLM’s embedding space. While our study focuses on factual recall, future work should examine other tasks, comparing VLMs against their LLM backbone.

Overall, this study provides strong empirical evidence that representation misalignment limits VLMs' ability to re-use the existing task circuits in their LLM backbone models. Further investigating and mitigating this issue is crucial for improving multimodal reasoning. 

\section*{Contributions}
Constantin Venhoff conceived and led the project, designed the methodology, and conducted all experiments. Ashkan Khakzar and Sonia Joseph provided in-depth feedback throughout and contributed to the manuscript. Philip Torr and Neel Nanda advised on the research and provided high-level feedback.

\section*{Acknowledgements}
This work was carried out as part of the ML Alignment \& Theory Scholars (MATS) program. We thank Josh Engels, Julian Minder, Clément Dumas, Iván Arcuschin, Nick Jiang and Sharan Maiya for helpful comments, discussions and feedback.

%% file: sec/X_suppl.tex
\clearpage
\setcounter{page}{1}

\section{Compute Resources}
We used an NVIDIA A100 GPU with 80GB memory for all local experiments. Factual question generation for the benchmark was performed via API calls to GPT-4.1.

\section{Dataset Generation Pipeline}
\label{sec:dataset-pipeline}

\begin{figure}[ht]
    \centering
    \includegraphics[width=\linewidth]{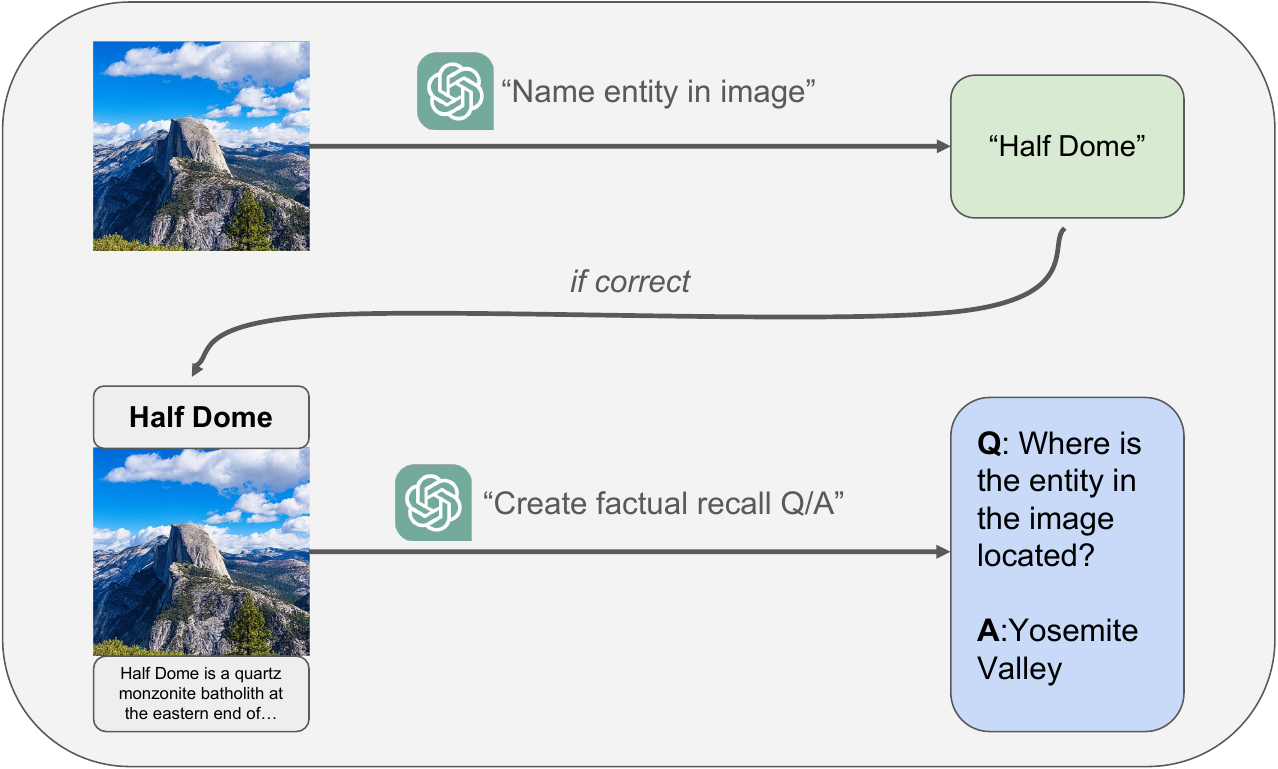}
    \caption{
Overview of the dataset construction pipeline for the multimodal factual recall benchmark.
}
\label{fig:dataset-pipeline}
\end{figure}

We illustrate the pipeline to generate factual recall questions in Figure \ref{fig:dataset-pipeline}. The procedure has five core stages:
\begin{itemize}
    \item  \textbf{Image Sampling:} Images are randomly selected from WIT and paired with their ground-truth entity names from the page title;
    \item \textbf{Entity Detection:} GPT-4.1 is prompted to list the central identifiable entity in the image;
    \item \textbf{Entity Verification:} GPT-4.1 checks whether any detected entity matches (or paraphrases) the ground-truth entity name;
    \item \textbf{Question Generation:} Given the entity name and context from the Wikipedia page, a factual recall question about the entity is generated, without naming the entity explicitly;
    \item \textbf{Paraphrase Generation:} To minimize API calls during the benchmark run, we sample a list of common paraphrases for the entity name, enabling us to use simple string matching for entity guesses.
    \item \textbf{Save to Dataset:} The image, verified entity name (and paraphrases), and generated question are stored as a finalized benchmark sample.
\end{itemize}

\section{Token Space Misalignment}

\begin{figure}[h]
\centering
\includegraphics[width=\linewidth]{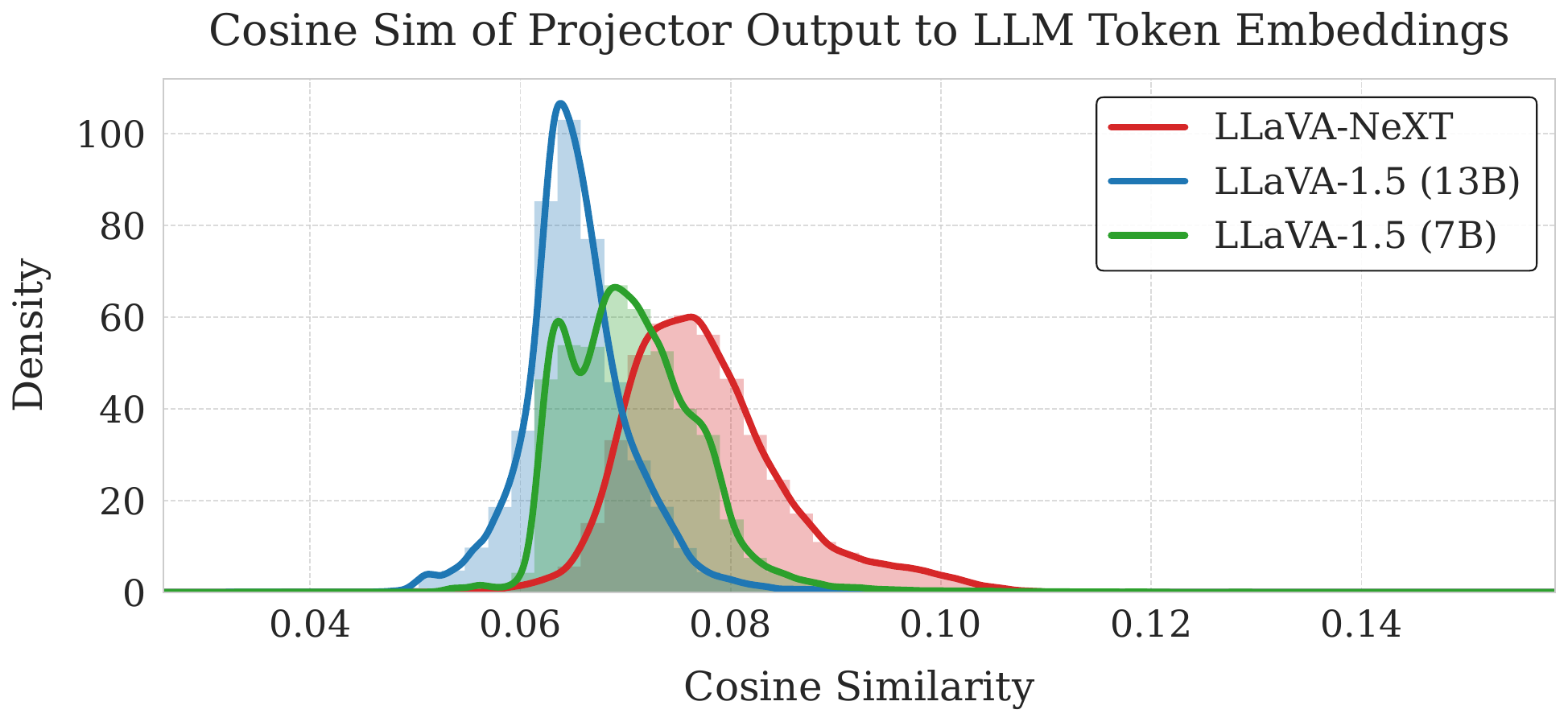}
\caption{Distribution of absolute cosine similarity between visual projector outputs and text embeddings for LLaVA-1.5-7B, LLaVA-1.5-13B, and LLaVA-MORE. All models exhibit similar distributions, with means ranging from approximately 0.06 to 0.08. The distributions are characterized by their sharp, concentrated profiles with high density around their respective means. This concentrated distribution pattern strongly indicates that visual token embeddings consistently occupy subspaces nearly orthogonal to the pretrained text token embedding space across all evaluated models.}
\label{fig:exp1}
\end{figure}

\label{app:token_space}
We reproduce the results of prior work \citep{masry2025alignvlm}, which shows that projector outputs of LLaVA-style VLMs are not aligned with the token space of their backbone LLMs, using LLaVA-1.5-7B, LLaVA-1.5-13B, and LLaVA-MORE. We extract visual projector outputs from 1,000 randomly selected ImageNet images. Each image produces 575 visual token activations, yielding a total of 575,000 activations per model. To assess alignment, we randomly sample 10,000 visual token embeddings per model and compute their cosine similarity with textual token embeddings from the LLM backbone.

\paragraph{Results and Analysis:}
Figure~\ref{fig:exp1} presents the distribution of cosine similarity scores between visual projector outputs and textual embeddings. The results indicate a pronounced misalignment across all evaluated models, with cosine similarity scores tightly concentrated near zero. This suggests that visual tokens predominantly occupy an embedding subspace orthogonal to pretrained textual representations.

These findings provide direct empirical support for our hypothesis that visual embeddings fail to integrate naturally into the LLM backbone’s structured token space. As a result, early-layer factual recall mechanisms remain disengaged when processing visual inputs, reinforcing the idea that factual recall degradation in VLMs stems from fundamental adapter misalignment rather than insufficient computational depth alone.

\section{Attribution Patching Noise Multiplier}
\label{app:noise}

\begin{figure}[h]
\centering
\includegraphics[width=\linewidth]{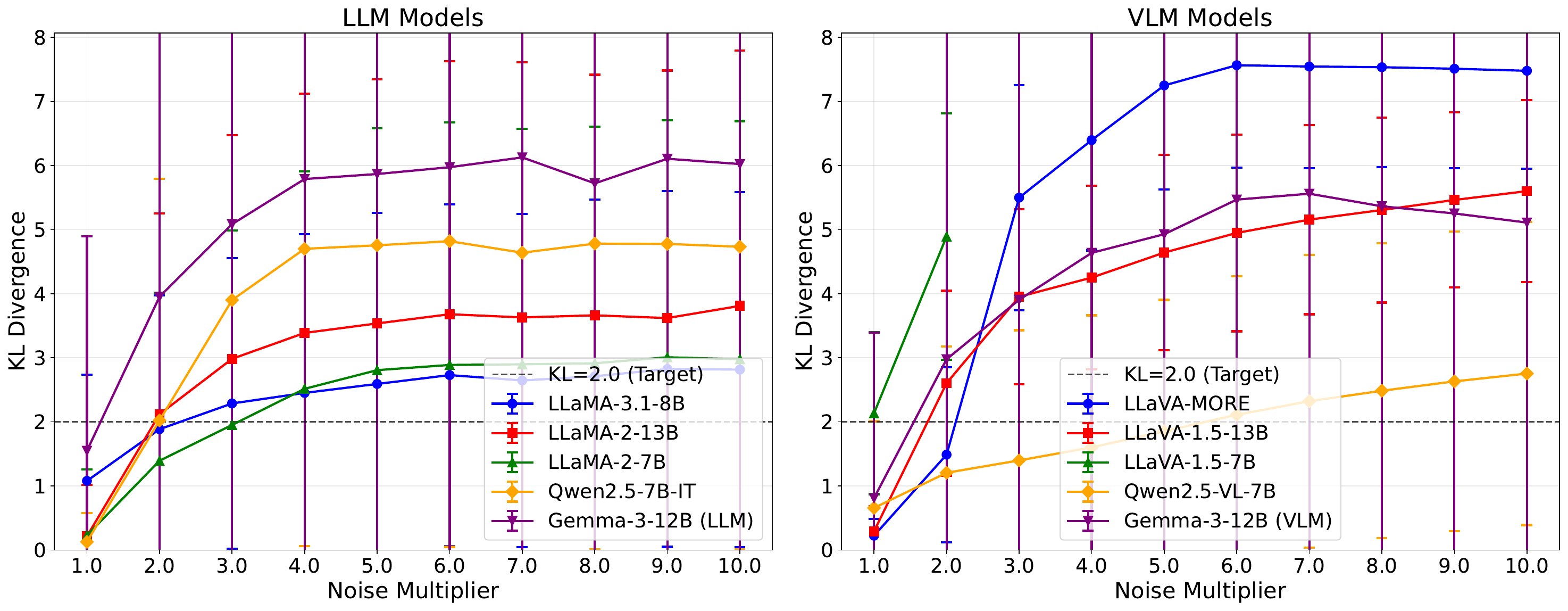}
\caption{The x-axis shows different choices for the noise multiplier $\alpha$. The y-axis shows the average KL Divergence of the clean predicted token distribution and the corrupted predicted token distribution, using the indicated $\alpha$ for the corruption. (For LLaVA-1.5-7B we only show values for $\alpha \in \{1,2\}$ as other values yielded NaN for the KL Divergence.)}
\label{fig:noise_multiplier}
\end{figure}

The noise multiplier determines how strongly we corrupt the entity input embeddings (text tokens or image tokens). We aim for a choice that leads the model to predict a false answer, without completely corrupting its generated output. By experimenting with different values we find that a KL Divergence around $2$ usually works well for the tested models. Therefore, we ablate $\alpha$ for each model and choose the value that causes a KL Divergence closest to $2$. Results are shown in \ref{fig:noise_multiplier}.

For LLaVA-1.5-7B we choose 1, for LLaVA-1.5-13B we choose 2, for LLaVA-More we choose 2, for Qwen2.5-7B-Instruct we choose 2, for Gemma-3-12B (LLM) we choose 1, for Qwen2.5-VL-7B-Instruct we choose 6, and for Gemma-3-12B (VLM) we choose 2.

\section{Layer Selection for Heuristic Patching}
\label{app:patching}
\begin{figure}[h]
\centering
\includegraphics[width=\linewidth]{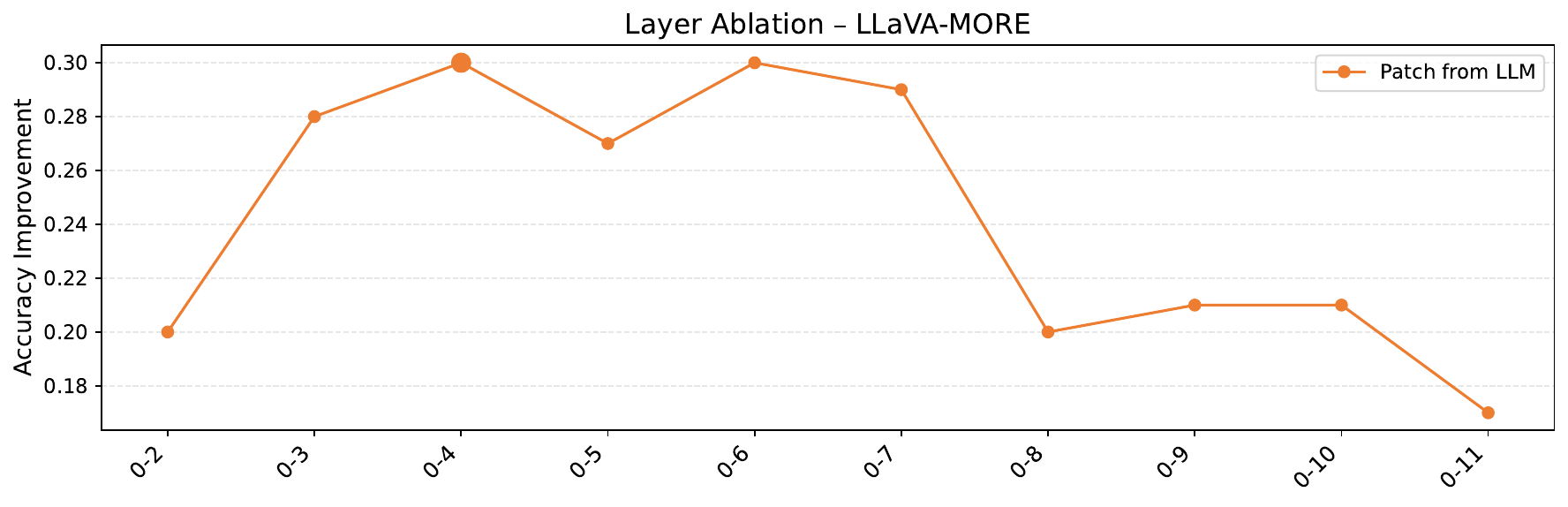}
\caption{The x-axis shows the layer range used to cache and patch MLP activations from the original LLM backbone model into the according LLaVA-style VLM. The y-axis shows the recovered performance.}
\label{fig:llava_more_patching}
\end{figure}
\begin{figure}[h]
\centering
\includegraphics[width=\linewidth]{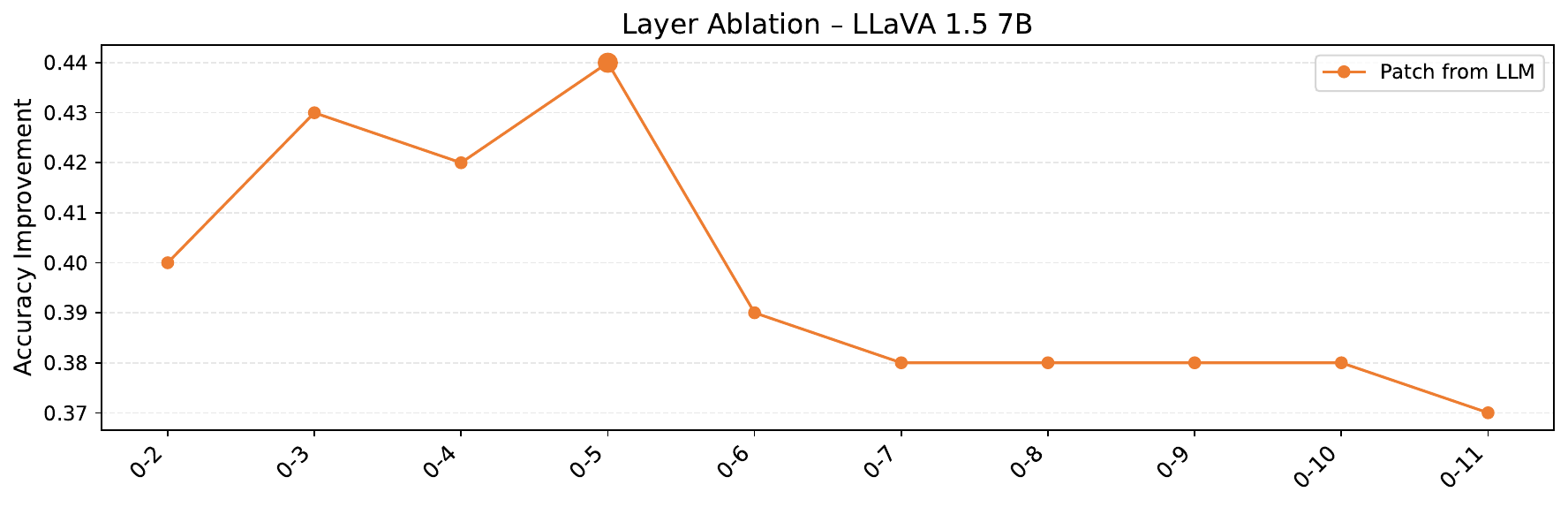}
\caption{The x-axis shows the layer range used to cache and patch MLP activations from the original LLM backbone model into the according LLaVA-style VLM. The y-axis shows the recovered performance.}
\label{fig:llava_7b_patching}
\end{figure}
\begin{figure}[h]
\centering
\includegraphics[width=\linewidth]{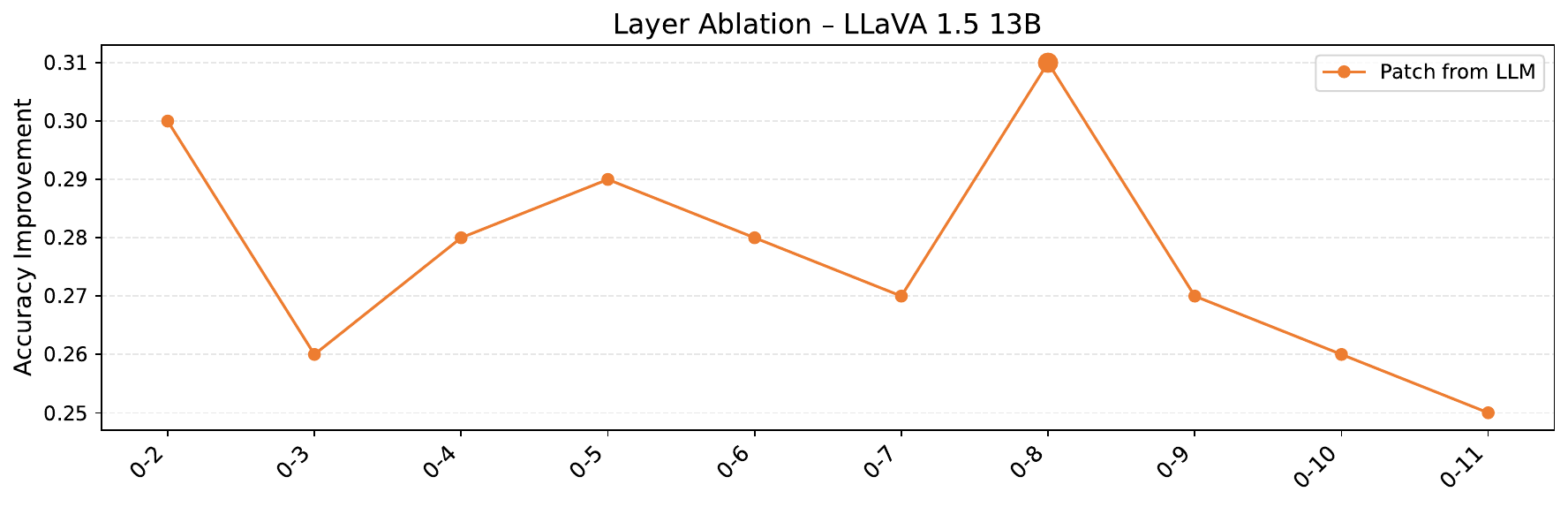}
\caption{The x-axis shows the layer range used to cache and patch MLP activations from the original LLM backbone model into the according LLaVA-style VLM. The y-axis shows the recovered performance.}
\label{fig:llava_13b_patching}
\end{figure}

We run heuristic patching across several layer ranges, to capture the impact of the layer choice. For the final plot we use the best performing layer range. We see across all models a relatively consistent maximum, with a decrease in accuracy towards larger layer ranges, indicating that the early-layer MLPs are indeed causally most relevant for factual recall. Results for LLaVA-MORE are shown in \ref{fig:llava_more_patching}, for LLaVA-1.5-7B in \ref{fig:llava_7b_patching}, and for LLaVA-1.5-13B in \ref{fig:llava_13b_patching}.

\section{Benchmark Prompt Templates}
\label{app:prompts}

The placeholders \texttt{\{entity\_name\}} and \texttt{\{question\}} are programmatically substituted at runtime.

\subsection{Image entity recognition (VLM)}
\begin{verbatim}
I provide you an image from wikipedia. Can you tell me what the specific name of the entity shown in the image is?
The specific name is the name the entity is commonly known by (name of person e.g. Julius Caesar, name of location e.g. Eiffel Tower, name of building e.g. White House, etc.).

Required JSON fields:
- answer: A string containing the specific entity name

Example 1:
Input: Image of the Eiffel Tower in Paris
Output: {{
    "answer": "Eiffel Tower"
}}

Example 2:
Input: Image of a famous scientist
Output: {{
    "answer": "Albert Einstein"
}}

Answer in the following json format: 
{{
    "answer": "<specific_entity_name>"
}}
\end{verbatim}

\subsection{LLM factual recall (base)}
\begin{verbatim}
Consider the following entity: {entity_name}.
Answer the following question: {question}

Required JSON fields:
- answer: A string containing the final answer in 1-5 words

Example 1:
Input: Entity: Eiffel Tower, Question: When was it completed?
Output: {{
    "answer": "1889"
}}

Example 2:
Input: Entity: Albert Einstein, Question: What theory is he most famous for?
Output: {{
    "answer": "Theory of Relativity"
}}

PROVIDE YOUR FINAL ANSWER IN 1-5 WORDS IN THE FOLLOWING FORMAT AND PROVIDE NO OTHER TEXT:
{{
    "answer": "<your answer>"
}}
\end{verbatim}

\subsection{VLM factual recall (base)}
\begin{verbatim}
I provide you with an image of an entity.
Answer the following question: {question}

Required JSON fields:
- answer: A string containing the final answer in 1-5 words

Example 1:
Input: Image of the Eiffel Tower, Question: When was it completed?
Output: {{
    "answer": "1889"
}}

Example 2:
Input: Image of Albert Einstein, Question: What theory is he most famous for?
Output: {{
    "answer": "Theory of Relativity"
}}

PROVIDE YOUR FINAL ANSWER IN 1-5 WORDS IN THE FOLLOWING FORMAT AND PROVIDE NO OTHER TEXT:
{{
    "answer": "<your answer>"
}}
\end{verbatim}

\subsection{LLM factual recall (CoT)}
\begin{verbatim}
Consider the following entity: {entity_name}.
Answer the following question: {question}
Reason through this question by:
1. Summarizing relevant background knowledge about the entity
2. Deriving the answer to the question

Required JSON fields:
- reasoning: A string containing 2-3 sentences explaining your thought process
- answer: A string containing the final answer in 1-5 words

Example 1:
Input: Entity: Eiffel Tower, Question: When was it completed?
Output: {{
    "reasoning": "The Eiffel Tower was built for the 1889 World's Fair in Paris. Construction began in 1887 and took about 2 years to complete.",
    "answer": "1889"
}}

Example 2:
Input: Entity: Albert Einstein, Question: What theory is he most famous for?
Output: {{
    "reasoning": "Einstein revolutionized physics with his theory of relativity. His most groundbreaking work was the theory of general relativity, which he published in 1915.",
    "answer": "Theory of Relativity"
}}

PROVIDE YOUR REASONING OF 2-3 SENTENCES AND FINAL ANSWER IN of 1-5 WORDS IN THE FOLLOWING FORMAT AND PROVIDE NO OTHER TEXT:
{{
    "reasoning": "<your reasoning>",
    "answer": "<your answer>"
}}
\end{verbatim}

\subsection{VLM factual recall (CoT)}
\begin{verbatim}
I provide you with an image of an entity.
Answer the following question: {question}
Reason through this question by:
1. providing a brief description of the image
2. Identify the entity the question refers to and summarize relevant background knowledge about the entity
3. Derive the answer to the question

Required JSON fields:
- reasoning: A string containing 3-4 sentences explaining your thought process
- answer: A string containing the final answer in 1-5 words

Example 1:
Input: Entity: Eiffel Tower, Question: When was it completed?
Output: {{
    "reasoning": "This is an image of the Eiffel Tower in Paris, a wrought-iron lattice tower. It was built for the 1889 World's Fair and took about 2 years to construct, with completion in 1889.",
    "answer": "1889"
}}

Example 2:
Input: Image of Albert Einstein, Question: What theory is he most famous for?
Output: {{
    "reasoning": "This is an image of Albert Einstein, a renowned physicist known for his revolutionary contributions to physics. His most groundbreaking work was the theory of general relativity, which he published in 1915 and fundamentally changed our understanding of space and time.",
    "answer": "Theory of Relativity"
}}

PROVIDE YOUR REASONING OF 3-4 SENTENCES AND FINAL ANSWER IN of 1-5 WORDS IN THE FOLLOWING FORMAT AND PROVIDE NO OTHER TEXT:
{{
    "reasoning": "<your reasoning>",
    "answer": "<your answer>"
}}
\end{verbatim}

\section{Case Study into Excluded Images}
\label{app:exclusion}
We conduct a brief case study into the images for which entities are not correctly identified by VLMs. We check the entity recognition results of Pixtral-12B, and find that about 24\% of examples are excluded due to Pixtral-12B not identifying the correct entity. A manual audit of 100 rejected cases reveals four overall failure modes:

\begin{enumerate}
    \item \textbf{Paraphrase mismatch} – e.g., ``FotoArtFestival'' $\rightarrow$ ``6 Foto Art Festival''
    \item \textbf{Near-entity confusion} – e.g., ``SS Monterey'' $\rightarrow$ ``RMS Mauretania''
    \item \textbf{Imprecision} – e.g., ``City Point, Wisconsin'' $\rightarrow$ ``Jackson County''
    \item \textbf{Completely off-target} – rare; e.g., ``Grand Palace Hotel'' $\rightarrow$ ``European Union''
\end{enumerate}

We find that paraphrase mismatches and imprecision occur most often, likely because the VLM genuinely does not know the correct entity. Paraphrase mismatches in general are infrequent, which supports our approach of pre-generating a list of paraphrases to avoid an API call for every entity guess. Lastly, we find very few ``completely off-target'' cases. These seem to arise when the VLM misidentifies the central entity, e.g., the image above contains a European Union flag on the hotel, and the VLM assumes the target entity is the European Union rather than the hotel. No systematic bias toward any particular domain (e.g., geography, people, or protected classes) was observed. Images span landscapes, historical objects, banknotes, architecture, etc.; the error distribution looks random rather than skewed. Consequently, we believe these exclusions are unproblematic.